\documentclass[10pt,letterpaper]{article}

\usepackage{multirow}
\usepackage{subfig}

\usepackage[top=0.85in,left=2.75in,footskip=0.75in]{geometry}

\usepackage{amsmath,amssymb}

\usepackage{changepage}

\usepackage[utf8x]{inputenc}

\usepackage{textcomp,marvosym}

\usepackage{cite}

\usepackage{nameref,hyperref}

\usepackage[right]{lineno}

\usepackage{microtype}
\DisableLigatures[f]{encoding = *, family = * }

\usepackage[table]{xcolor}

\usepackage{array}

\newcolumntype{+}{!{\vrule width 2pt}}

\usepackage{float}

\newlength\savedwidth

\newcommand\thickhline{\noalign{\global\savedwidth\arrayrulewidth\global\arrayrulewidth 2pt}%
\hline
\noalign{\global\arrayrulewidth\savedwidth}}

\raggedright
\usepackage[aboveskip=1pt,labelfont=bf,labelsep=period,justification=raggedright,singlelinecheck=off]{caption}

\makeatletter
\renewcommand{\@biblabel}[1]{\quad#1.}
\makeatother

\usepackage{lastpage,fancyhdr,graphicx}
\usepackage{epstopdf}
\fancyheadoffset[L]{2.25in}
\fancyfootoffset[L]{2.25in}
\begin{document}

\vspace*{0.2in}

\begin{flushleft}
{\Large
\textbf\newline{Automatic segmentation of the Foveal Avascular Zone in ophthalmological OCT-A images} 
}
\newline
\\
Macarena D\'iaz\textsuperscript{1,2\Yinyang*},
Jorge Novo\textsuperscript{1,2\Yinyang},
Paula Cutr\'in\textsuperscript{3\ddag},
Francisco G\'omez-Ulla\textsuperscript{3,4\ddag},
Manuel G. Penedo\textsuperscript{1,2\Yinyang},
Marcos Ortega\textsuperscript{1,2\Yinyang},
\\
\bigskip
\textbf{1} Department of Computer Science, University of A Coruña, A Coruña (Spain)
\\
\textbf{2} CITIC-Research Center of Information and Communication Technologies,
    	University of A Coruña, A Coruña (Spain)
\\
\textbf{3} Complejo Hospitalario Unversitario de Santiago, Santiago de Compostela (Spain)
\\
\textbf{4} Instituto Oftalmol\'ogico G\'omez-Ulla, Santiago de Compostela (Spain)
\\
\bigskip

%
%
\Yinyang These authors contributed equally to this work.

\ddag These authors also contributed equally to this work.




* macarena.diaz1@udc.es

\end{flushleft}
\section*{Abstract}
Angiography by Optical Coherence Tomography (OCT-A) is a non-invasive retinal imaging modality of recent appearance that allows the visualization of the vascular structure at predefined depths based on the detection of the blood movement through the retinal vasculature. In this way, OCT-A images constitute a suitable scenario to analyse the retinal vascular properties of regions of interest as is the case of the macular area, measuring the characteristics of the foveal vascular and avascular zones. Extracted parameters of this region can be used as prognostic factors that determine if the patient suffers from certain pathologies (such as diabetic retinopathy or retinal vein occlusion, among others), indicating the associated pathological degree. The manual extraction of these biomedical parameters is a long, tedious and subjective process, introducing a significant intra and inter-expert variability, which penalizes the utility of the measurements. In addition, the absence of tools that automatically facilitate these calculations encourages the creation of computer-aided diagnosis frameworks that ease the doctor's work, increasing their productivity and making viable the use of this type of vascular biomarkers.

In this work we propose a fully automatic system that identifies and precisely segments the region of the foveal avascular zone (FAZ) using a novel ophthalmological image modality as is OCT-A. The system combines different image processing techniques to firstly identify the region where the FAZ is contained and, secondly, proceed with the extraction of its precise contour. The system was validated using a representative set of 168 OCT-A images, providing accurate results with the best correlation with the manual measurements of two experts clinician of 0.93 as well as a Jaccard's index of 0.82 of the best experimental case. This tool provides an accurate FAZ measurement with the desired objectivity and reproducibility, being very useful for the analysis of relevant vascular diseases through the study of the retinal microcirculation.



\section*{Introduction}

Over the recent years, the constant technological advances allow the integration of specialized computed-aided diagnosis systems in different fields of medicine \cite{novo2017hydra,novo2017wivern,po007}. These systems ease the doctor's work, facilitating and accelerating the diagnosis and monitoring of many diseases, in addition to the inclusion of important advantages as objectivity and determinism that are not always present in the diagnostic processes of the experts in their clinical routine. These facts are present in ophthalmology, where the analysis and diagnostic procedures frequently involve the use of different image modalities as a relevant source of information of a large variability of relevant diseases. Among the ophthalmological image modalities, in the recent years, we can find the appearance of the Angiography by Optical Coherence Tomography (OCT-A) that is a new non-invasive imaging modality that allows the visualization, with great precision, of the vasculature at different depths over the retinal eye fundus. OCT-A images are mainly based on the detection of blood movement without the need of injecting intravenous contrast, fact that was unavoidable in previous capture techniques, as happens with classic angiographies. The classic angiography is a simple but invasive image modality that allows the study of the vascular characteristics of the retina using the injection of an intravenous contrast to the patient. Subsequently, Optical Coherence Tomography (OCT) \cite{de2017enhanced} allows to observe, non-invasively, a cross-sectional visualization of the layers of the retina. Finally, OCT-A combines the advantages of both, offering a suitable visualization for the analysis of the retinal vasculature, as angiographies, but non-invasively, using the tomography capture characteristics, which constitutes a more comfortable scenario for the patients. OCT-A images are typically taken at superficial and deep views of the eye fundus, which facilitates the subsequent vascular analysis; in addition, these images can be obtained at different levels of zoom, being 3 and 6 millimeters-wide (greater and smaller zooms) the most used configurations. This image technique offers many advantages \cite{octa-review} compared to those previously used, such as the possibility of generating volumetric scans that can be captured at specific depths, offering a 3D visualization of the eye fundus with a limited time and cost that it typically involves (image acquisition in about 2 or 3 seconds). Given these characteristics, OCT-A images are suitable for the analysis of the retinal micro-circulation, being spread their use in many health-care systems.

The higher or lower presence of vessels in certain areas of the eye fundus is a very useful biomedical parameter since they are affected by many vascular pathologies, such as diabetic retinopathy or age related macular degeneration, being their level of presence or absence a significant prognostic factor. One of these parameters is the area of the Foveal Avascular Zone (FAZ), the region of the fovea that has no blood supply. The analysis of the FAZ region is crucial given its characteristics are directly related to many relevant clinical conditions. As reference, it is related to the visual acuity of patients who suffer from diabetic retinopathy or the occlusion of the retinal vein \cite{visual-acuity}.

As reference, the population with diabetes has from 40\% to 90\% of suffering from diabetic retinopathy; in addition, people with diabetic retinopathy are 5 times more likely to derive in total blindness. Given those facts, the identification, segmentation and analysis of the FAZ region is crucial for the early diagnosis of relevant diseases as diabetic retinopathy.

Given that it is a recently technology, there are still few studies that are related to the automatic extraction of measurements of interest using the OCT-A image modality.
Instead, these early studies are mainly based on the clinical analysis of these images to define manual parameters that can be extracted and the characteristics they typically offer \cite{po01} .
There are works that study the repeatability and reproducibility of these measures in healthy patients \cite{interesdos,interes} indicating the satisfactory impact of this analysis. In addition, as previously indicated, it was shown that visual acuity is related to the FAZ area in patients with diabetic retinopathy and with the occlusion of the retinal vein \cite{visual-acuity}, demonstrating the suitability and the clinical relevance of this analysis in the diagnosis of relevant pathologies related to the vision loss.
However, still few proposed computational studies are based on the extraction of the FAZ region. Lu \textit{et al.} \cite{lu2018evaluation} faces the automatic FAZ extraction and its quantification in different measurements to classify the images as healthy or diabetic cases. Particularly, the FAZ region is extracted applying a region growing approach in the exact central point of the image as seed, which represents a significative limitation with the initialization of this static point; then, morphological operators and an active contour model are applied in order to obtain the final FAZ segmentation. Next, four different parameters are calculated to quantify the FAZ region and classify the image as a healthy or diabetic case. In the work of Hwang \textit{et al.} \cite{article}, the proposal directly subtracts the image intensities over consecutive OCT-A images in order to generally obtain avascular zones, deleting posteriorly the non-representatives ones using a given size as reference.

In this paper, we propose a fully automated and robust methodology to localized and measure the FAZ region in OCT-A images. The validation of the proposal was performed with a set of experiments, using a representative a public dataset that covered a significative age-range as well as modalities of healthy and diabetic OCT-A images. Specifically, this public dataset contains $3 \times 3$ millimeters and $6 \times 6$ millimeters superficial and depth healthy OCT-A images from people between 10 and 69 years old, including all the types in each age-range. Moreover, a smaller part of the dataset belongs to diabetic patients, including 6 images for each of the 4 mentioned subgroups: $3 \times 3$ millimeters superficial and depth and $6 \times 6$ millimeters superficial and depth. In the Section \textit{\nameref{sec:dataset}} we explain extensively these used image dataset.
The methodology that is presented in this work is able to perform the aforementioned actions automatically, without the need of the user intervention. Generally, the methodology to segment the FAZ region implies the following steps: first, the image acquisition and normalization of its values in order to facilitate the following stages of the process; second, an exhaustive analysis of the image to detect FAZ candidates and the consecutive removal of existing false positives; then, from the remaining candidates, we select the correct FAZ; and, finally, a precise segmentation of the FAZ region is achieved. The obtained results were compared with the manual measurements of two expert clinicians to analyze the correlation and similarity of the results of the system with the manual performance of an expert clinician.

This paper is organized as follows: Section \textit{\nameref{sec:materials}} presents the OCT-A image dataset that was used in the experiments as well as the detailed characteristics of the proposed method. Section \textit{\nameref{sec:results}} exposes the results and comparisons with the manual segmentations. Finally, Section \textit{\nameref{sec:conclusions}} discusses about the obtained results, concludes the paper and indicates possible future lines of work.

\section*{Materials and methods}
\label{sec:materials}

\subsection*{Image dataset}
\label{sec:dataset}

The validation process was done using the public image dataset \textit{OCTAGON} \cite{octagon}, that contains 144 healthy and 24 diabetic OCT-A images, summing a total of 168 cases. The images were taken using the Optical Coherence Tomography capture device \textit{DRI OCT Triton; Topcon Corp} taking images from both left and right eyes of different patients. Additionally, the images were obtained at different levels of zoom and depths,with a resolution of $320 \times 320$ pixels. In particular, the following configurations were represented in the dataset:
\begin{itemize}
	\item Superficial. OCT-A images in which the foveal area can be observed from the surface.
	\item Deep. OCT-A images visualizing the deep foveal area.
\end{itemize}

The previous configurations were also captured at the following resolutions:

\begin{itemize}
	\item 3$\times$3 millimeters OCT-A images centered in the fovea covering a region of 3$\times$3 millimeters. Hence, a greater level of detail of the captured macular region is appreciated.
	\item 6$\times$6 millimeters  OCT-A images centered in the fovea covering a region of 6$\times$6 millimeters. Hence, a wider range of the macular region is visualized. 
\end{itemize}

\begin{figure}[H]
\centering
\includegraphics[width=120mm]{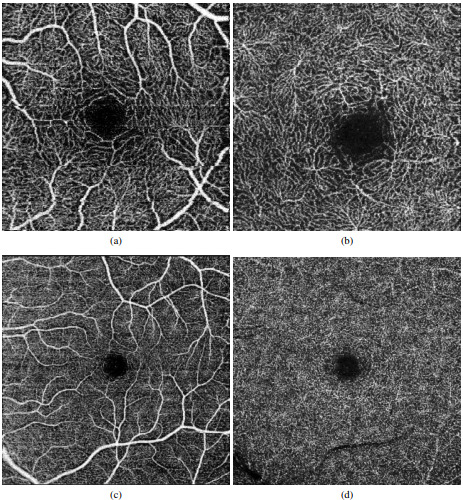}
	\caption {Examples of OCT-A images representing all the configurations that were used in this work. $1^{st}$ row, images of 3x3 millimeters. $2^{nd}$ row, images of 6x6 millimeters. (a) \& (c) Superficial OCT-A images. (b) \& (d), Deep OCT-A images.}
	\label{fig:ejemplos}
\end{figure}

Fig. \ref{fig:ejemplos} illustrates, with representative examples, all the 4 configurations that are represented in the used dataset. Additionally, the set of 144 healthy images presents the following clinical and population characteristics:

\begin{itemize}
	\item Age range. The image dataset is divided into 6 age ranges: 10-19 years, 20-29 years, 30-39 years, 40-49 years, 50-59 years and 60-69 years. This way, we used a diverse set of images with a significant variability of ages.
	\item Division by patients. For each mentioned age range, images from three different patients were captured.
	\item Eye. For each patient, we have OCT-A images that were extracted from both left and right eyes.
	\item Depth and size. Finally, for each eye, 4 images were captured ranging all the superficial/deep and 3/6 millimeters configurations.
\end{itemize}

Additionally, two expert clinicians manually labelled and segmented the FAZ region of each OCT-A image. This ground truth served as reference for the validation process of the method.

As said, the dataset also includes 24 diabetic OCT-A images, 6 of each mentioned subgroup. Given that these OCT-A images are manually labelled by an expert clinician, the validation process is the same as with healthy cases, testing that the method is valid for both healthy and diabetic OCT-A images.

\subsection*{Proposed methodology}

We based the proposed methodology on the analysis of the main image characteristics of the FAZ region as it typically appears in the OCT-A images. Generally, these characteristics are the following:
\begin{itemize}
	\item Macular centered area. Although this is not exactly in all the cases, the FAZ region is typically centered on the macular region, specially in the cases of healthy patients.
	\item Low intensity profile region. Given the absence of vasculature, the FAZ region is generally defined as a dark area with a significative contrast with respect to the neighbour areas of the macular region.
	\item Surrounded by blood vessels. Given this low intensity profile region, surrounded by blood vessels, we can base the precise delimitation of the FAZ region using this surrounding vasculature as reference.
\end{itemize}

\begin{figure}[H]
	\begin{center}
	\includegraphics[width=120mm]{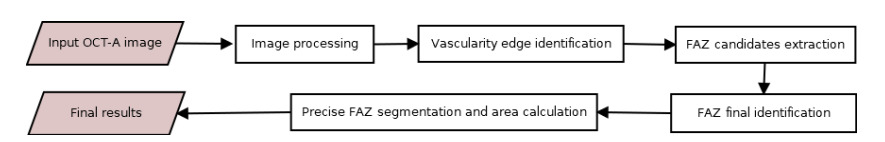}
		\caption{Main steps of the proposed methodology.}
		\label{fig:diagram}
	\end{center}
\end{figure}

The proposed methodology based its characteristics in these properties to achieve the desired results. Fig. \ref{fig:diagram} illustrates the main steps of the proposed method. They are progressively discussed in next subsections. 

\subsubsection*{Image processing}
\begin{figure}[H]
\centering
\includegraphics[width=120mm]{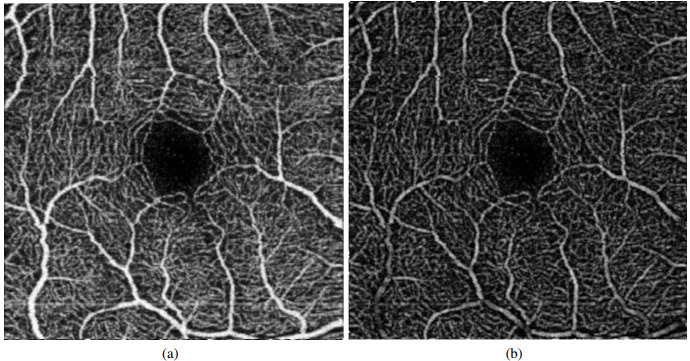}
	
	\caption{Application of the preprocessing step. (a) Original image. (b) Image result after applying the-top hat operator.}
	\label{fig:tophat_application}
\end{figure}

We initiality intensify the visualization of the vasculature to facilitate its following differentiation by applying morphological operators. Morphological operators are often used to highlight the geometric properties of the image. Our first purpose is to clearly differentiate what is an avascular zone and what is not, so the objective of the application of the morphological operators is to make this difference stronger.
Given its use in different works with satisfactory results \cite{pso09}, we apply the \textit {white top-hat} operator (see Fig. \ref{fig:tophat_application}), since it makes the bright areas of the image more intense. In this way, vessels will present higher intensities while areas without vessels will remain with low intensity profiles.

\subsubsection*{Vascular edge identification}
\begin{figure}[H]
\centering
\includegraphics[width=120mm]{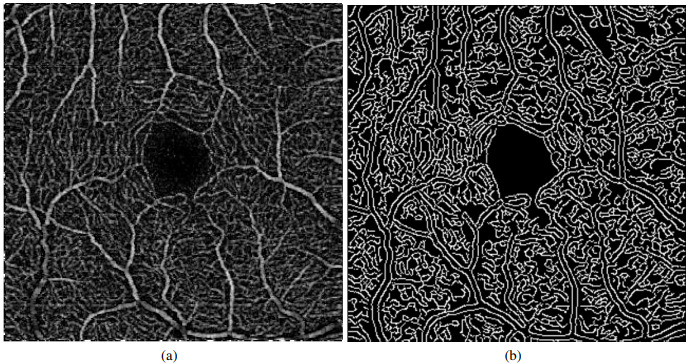}
	
	\caption{Vascularity edge identification using the Canny edge detector. (a) Original OCT-A image (after the top-hat preprocessing step). (b) Results of the vascular edge identification.}
	\label{fig:canny_metodologia}
\end{figure}

Using the previous image, we can easily identify the vascular regions and differentiate them from the target FAZ area. Additionally, this enhanced image also facilitates the removal of possible wrong identifications in subsequent stages of the methodology. To identify the vascularity, the Canny edge detector \cite{canny01} is used, extracting the edges of the vessels. 
The parameters of Canny edge detector are decisive for the results; in this case these parameters vary based on the image average values, allowing to acquire satisfactory results independently of the input OCT-A image.
This way, we obtain solid and continuous detections of the vasculature that serve as baseline for the vascular region identification. In Fig. \ref{fig:canny_metodologia}, we can see a representative example of the result after the application of the Canny edge detector.

\subsubsection*{Extraction of the FAZ candidates}
\begin{figure}[H]
\centering
\includegraphics[width=120mm]{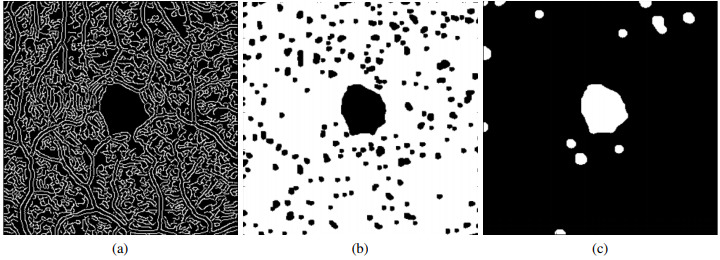}
	\caption{Morphological closure and inversion of intensities followed by a removal of small elements. (a) Image with the vascular edge identification. (b) Result after applying a morphological closure. (c) Result after applying an inversion of intensity and an opening.}
	\label{fig:open}
\end{figure}

Using the previous set of vascular detections as baseline, we identify all the regions that are suspicious of being candidates of the FAZ location. To remove most of the false positives, we firstly apply a morphological closure. The reason for choosing this operator instead of a dilatation is that the target vascular area would be excessively modified if an erosion is not applied after dilatation. Thus, after the application of the morphological closure we obtain an adequate scenario where we can easily identify the most suitable candidate as the target FAZ region, as illustrated in the example of Fig. \ref{fig:open}.

Afterwards, after inverting the image to facilitate posterior stages, an opening morphological operator is applied given that the previous image still contains a significant number of spurious detections. This way, as result, the fewer possible candidates (Fig. \ref{fig:open}(c)) are preserved.

\subsubsection*{FAZ region final identification}
\begin{figure}[H]
\centering
\includegraphics[width=120mm]{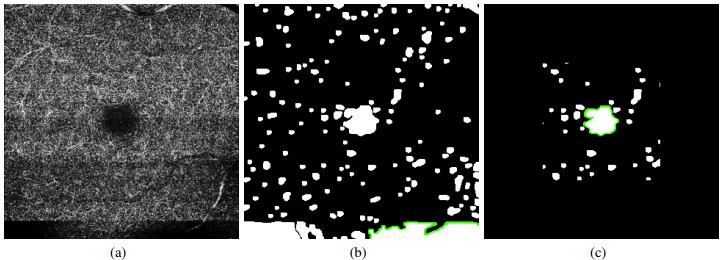}
  
	\caption{Example of error in the capture process. (a) Original image. (b) Initial set of identified FAZ candidates. (c) Final set of FAZ candidates after FP removal. }
	\label{fig:problem1}
\end{figure}

As indicated before, the main characteristics of the FAZ region imply a centered location, given that the OCT-A images are typically taken macular-centered, as well as their common appearance of low intensity profiles. These properties permitted that, in most of the cases, we obtain images from the previous stage as the case presented in Fig. \ref{fig:open}(b). In that cases, the larger identified region directly represents the FAZ region. However, other times, we face situations, as the one presented in Fig. \ref{fig:problem1}, where errors in the capture process or pathological conditions can introduce other significant dark regions in the OCT-A images, producing mistakes in the FAZ identifications. In that sense, we analysed the morphological characteristics of the remaining candidates to perform a precise identification, avoiding those that are not clearly FAZ regions. In particular, peripheral and disperse candidates are directly discarded and marked as background.

\begin{figure}[H]
\centering
\includegraphics[width=120mm]{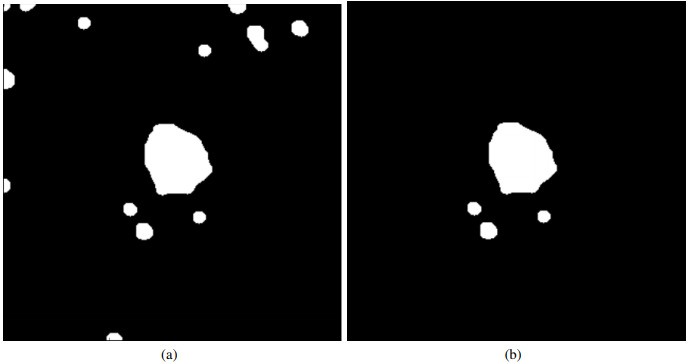}
  
	\caption{Removal process of FAZ FP candidates. (a) Initial set of identified FAZ candidates. (b) Final set of FAZ candidates.}
	\label{fig:im_original_sol}
\end{figure}

Applying these rules we can see that, as happens with the example of Fig. \ref{fig:problem1}(c), we remove many false positives, specially those problematic that could be confused with FAZ regions and therefore, introduce identification errors. Moreover, even without the existence of pathological or capture artefacts, this stage contributes discarding a significant number of FP candidates, as happens with the example of Fig. \ref{fig:im_original_sol}.

Finally, from the remaining candidates, we decide which of them represents the final FAZ identification. Carefully analysing the candidates, at this stage, we normally preserved the FAZ region and other small candidates of spurious artefacts. For that reason, we select the largest remaining candidate as the most significant one being, therefore, the identified FAZ region. There are many ways to check the largest sized regions. In our case, the used criterion is the measurement of the perimeter of the candidates. This preliminary extraction serves as baseline of the following precise FAZ segmentation.

\subsubsection*{Precise FAZ segmentation and area calculation}
The previously obtained FAZ segmentation is adequate in many cases. However, the use of morphological operators and the significant level of complexity of the OCT-A images penalize the segmentation precision in the surrounding FAZ limits. For that reason, we afterwards applied region growing \cite{537343,336259} using the previous segmentation as seed to further adjust with a higher precision the contour of the segmentation to the surrounding vascular edges. In this case, we implemented a new version of region growing, based on the original idea and adding new features. This implementation add to de original region growing the ability of deleting pixels that are in the region and not accomplish the region conditions.

Given that the preliminar segmentation could exceed the vasculature limits, we performed a preliminary erosion step to guarantee that the area that is used as seed for the region growing process is contained inside the real existing FAZ region. Then, the contour points of this seed are used by the region growing process to progressively aggregate or delete neighbouring pixels by intensity similarity until reaching the entire vascular edge contour. Finally, where no further pixels are added, the growing process is stopped.

\begin{figure}[H]
\centering
\includegraphics[width=120mm]{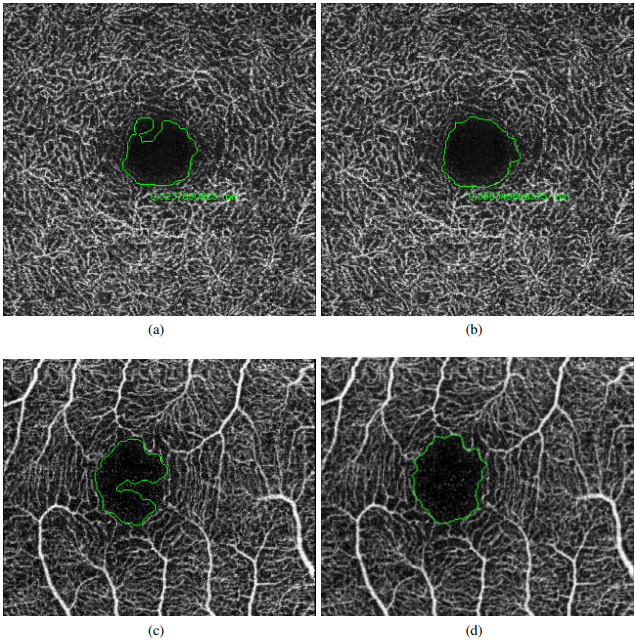}
	\caption{Application of the precise final FAZ segmentation. (a) \& (c) Preliminary FAZ extractions. (b) \& (d) Final segmentation results.}
	\label{fig:regiongrowing}
\end{figure}

The similarity criterion calculates the average intensity of the extracted region, letting a 30\% of variation as the tolerance for the addition of new pixels to the segmentation. This means that we accept a pixel in the region if it value is content in $[ARV - 0.3 \times ARV, ARV + 0.3 \times ARV]$, where \textit{ARV} is the average region value. Fig. \ref{fig:regiongrowing} presents a couple of imperfect preliminary FAZ extractions and their corresponding final precise segmentations. This way, we obtain more adjusted FAZ segmentations that are suitable for their use in following analyses and diagnostic processes.

Finally, using the resultant segmentation, the method also calculates the corresponding area of the identified FAZ zone, as a global and complementary numeric parameter to be used in clinical procedures. The area is calculated as follows:

\begin{equation}
	\label{areainmm}
	A = a \times \frac{mm^2}{height \times width}
\end{equation}

where \textit{a} represents the count of pixels of the segmented region, \textit{mm} represents the size in millimeters of the image (in our experiments 3 or 6 millimeters), and \textit{height} and \textit{width} indicates the dimensions of the analysed  OCT-A image.


\section*{Results}
\label{sec:results}

We conducted different experiments to validate the suitability of the proposed method using the image dataset that was presented in Section \textit{\nameref{sec:materials}}. As indicated, this dataset includes a significant variability of conditions with 144 images at superficial and deep levels as well as sizes of 3 and 6 millimeters. In the experiments, we compared the results of the method with the manual labelling of two experts clinician. The designed experiments were the following:

\begin{itemize}
\item Experiment 1. Validation of the accuracy of the localization process.
\item Experiment 2. Validation of the quality of the segmentation results. We performed a couple of comparisons: firstly, a global comparison analysing the area of the retrieved FAZ regions; secondly, a more adjusted comparison using the Jaccard's index.
\end{itemize}

Additionally, we divided the experiments by the analysis of the included 4 configurations of the OCT-A images, given the difference of complexity of each case. This way, we obtain more precise results and conclusions of the performance of the proposal in all the existing scenarios.

\subsection*{Experiment 1: validation of the FAZ localization stage}

We firstly tested if the proposal correctly identifies the location of the FAZ region that corresponds to the first part of the proposed methodology. This is a crucial stage as the subsequent precise FAZ segmentation depends on a preliminary correct detection. As gold standard, we consider that a localization was successfully achieved if the centroid of the preliminary extraction is placed inside the manual segmentation of the specialist.

\begin{table}[ht!]
\begin{adjustwidth}{-2.25in}{0in}
\caption{\bf Accuracy localization FAZ results using the proposed method in healthy OCT-A images.}
\label{tab:resultados_bien_mal}
\begin{center}
 \begin{tabular}{|l+l|l|l|}\hline
		\textbf{Size} & \textbf{Superficial} & \textbf{Deep} & \textbf{Total}\\ \thickhline
		\textbf{$3 \times 3$ millimeters} & 36/36 ($100\%$) & 36/36 ($100\%$) & 72/72 ($100\%$)\\ \hline
		\textbf{$6  \times 6$ millimeters} & 32/36 ($88.89\%$) & 36/36 ($100\%$) & 68/72 ($94.45\%$)\\ \hline
 \end{tabular}
\end{center}
\end{adjustwidth}
\end{table}

\begin{table}[ht!]
\begin{adjustwidth}{-2.25in}{0in}
\caption{\bf Accuracy localization FAZ results using the proposed method in diabetic OCT-A images.}
\label{tab:resultados_bien_mal_diabetic}
\begin{center}
 \begin{tabular}{|l+l|l|l|}\hline
		\textbf{Size} & \textbf{Superficial} & \textbf{Deep} & \textbf{Total}\\ \thickhline
		\textbf{$3 \times 3$ millimeters} & 6/6($100\%$) & 6/6 ($100\%$) & 12/12 ($100\%$)\\ \hline
		\textbf{$6  \times 6$ millimeters} & 6/6($100\%$) & 6/6 ($100\%$) & 12/12 ($100\%$)\\ \hline
 \end{tabular}
\end{center}
\end{adjustwidth}
\end{table}

\begin{figure}[H]
\centering
\includegraphics[width=120mm]{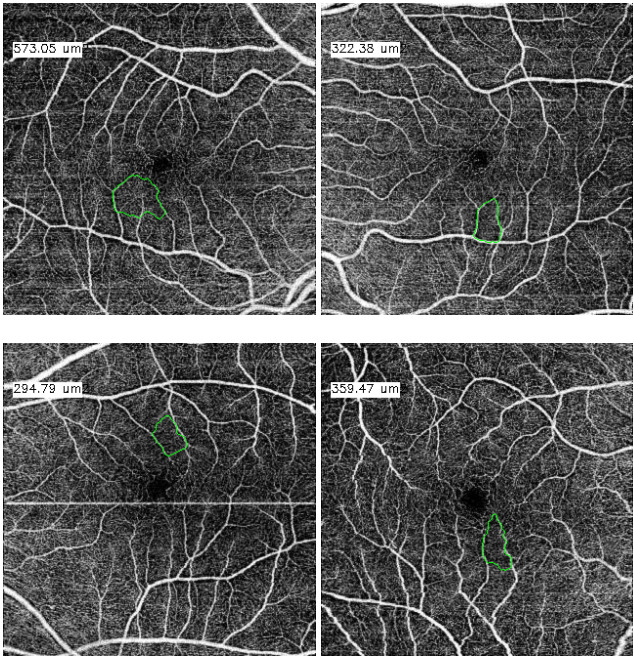}
	\caption{Error cases on the FAZ localization process.}
	\label{fig:errores}
\end{figure}

Table \ref{tab:resultados_bien_mal} summarizes the main localization results including the success and failure rates for both superficial and deep OCT-A healthy images. As we can see, the results using deep images were satisfactory, localizing correctly all the aimed 72 FAZ cases. Regarding the superficial images, the method also provided accurate results in most of the cases, remaining 4 cases where it was not correctly detected (the 4 cases are presented in Fig. \ref{fig:errores}). About these cases, they belong to 6 millimeters images, where the tonalities of the images are fairly regular and the FAZ normally presents small dimensions. This short size can make that the final selection of the biggest candidate returns a candidate that does not belong to the real FAZ region, discarding the real detected one. Despite that, we would like to highlight that this situation is only present in a very low number of particular cases.

Table \ref{tab:resultados_bien_mal_diabetic} summarizes the localization results and the success and failure rates for both superficial and deep OCT-A diabetic images, reaching accurate results in all the subgroups.

\subsection*{Experiment 2: validation of the FAZ segmentation}
\label{e2}
Over the correctly localized FAZ regions, we further analysed the characteristics of the obtained FAZ precise segmentations in comparison with the manual segmentations of the specialist. We firstly compared the final area size of the extracted regions given this is the aimed final parameter that is being used by clinicians in the diagnostic procedures, providing a general and bright idea about the usefulness of the results for their final purpose. In particular, we used the correlation coefficient \cite{doi:10.1177/875647939000600106} to measure the similarity and relationship between both method and clinician extracted area sizes. In this way, it can be verified whether the relationship between both sets is directly proportional. More formally, the correlation is calculated as the quotient between the covariance and the product of the standard deviations of both area size sets:

\begin {equation}
	\begin{split}
		&r = \frac{\sigma_{xy}}{\sigma_{x} \times \sigma_{y}}\\ 
	\end{split}
\end {equation}

The results of this operation returns values in the interval
 $[-1,1]$, where:
\begin{itemize}
\item $1 \ge r > 0$. The correlation between both sets is directly proportional, being r = 1 the maximum possible correlation.
\item r = 0. No correlation is identified between both sets.
\item $0 > r \ge -1$. The correlation is inversely proportional, being r = -1 the maximum inverse correlation.
\end{itemize}

\begin{figure}[H]
\centering
\includegraphics[width=120mm]{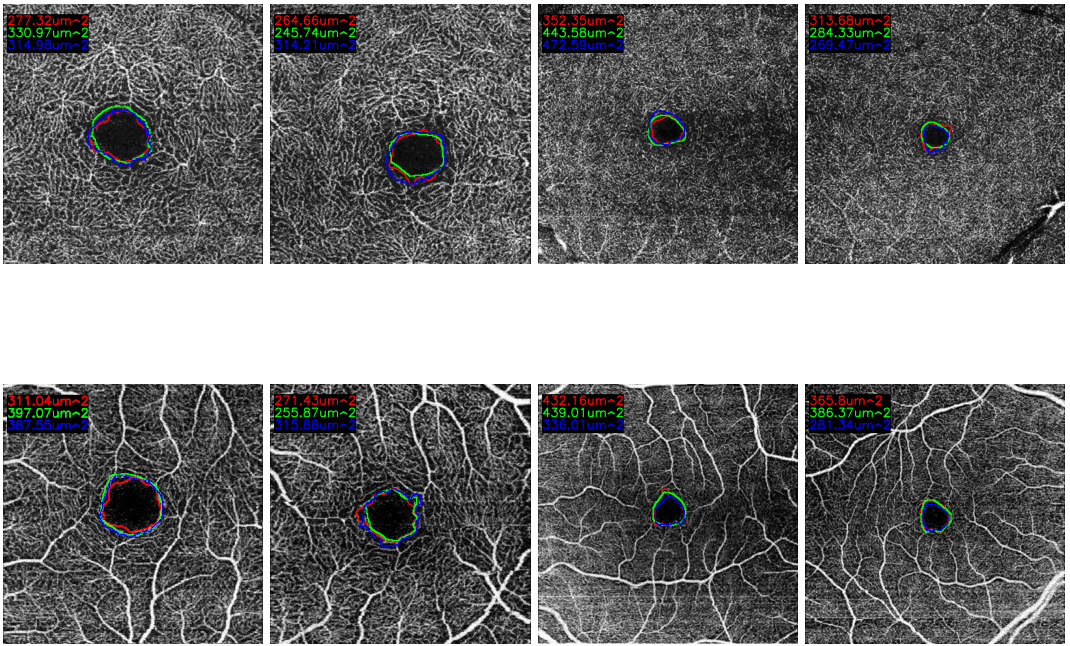}
  
	\caption{Comparative examples of the experts (green and red) and the automatic computational (blue) segmentations as well as the corresponding area size measurements.}
	\label{fig:comparativa}
\end{figure}

The results of the correlation are presented in  Table \ref{tab:resultados}, where we can observe that the performance of the proposed method is significantly correlated to the manual performance of the specialists, being even clearer this similarity in the cases of superficial OCT-A images. The higher values in superficial images are obtained given that the FAZ regions in these images present clearer surrounding vascular edges, for what the manual and the automatic region identifications agree with higher rates. Also, we have to consider the typical variability and imperfection of the manual identifications that are normally made by the specialists, instead of the determinism and repeatability of the computational performance of our proposal, which puts in valuable consideration the correlation rates that were obtained. In addition, it should be noted that the correlation between the specialists do not reach the highest value of the correlation coefficient of Pearson.This means that there is a discrepancy between the performance of both experts. Consequently, the correlation of the automatic system performance and the expert results is also penalized.  Fig. \ref{fig:comparativa} presents representative examples of superficial and deep images with the manual and automatic FAZ extractions and the area size measurements. As we can see, the similarity in the measurements motivates the significative results that were presented in Table \ref{tab:resultados}.

\begin{table}[ht!]
\begin{adjustwidth}{-2.25in}{0in}
\caption{\bf Correlation coefficients that were obtained using the manual and the automatic area size measurements in healthy OCT-A images.}
\label{tab:resultados}
\begin{center}
 \begin{tabular}{|l+l|l|l|}\hline
		\textbf{Size} & \textbf{Superficial} & \textbf{Deep} & \textbf{Comparisons}\\
		\thickhline
		\multirow{3}{*}{\textbf{3 $\times$ 3 millimeters}} & 0.93 & 0.54 & Expert1 vs. Expert2\\
																							& 0.90 & 0.81 & System vs. Expert2\\
																							& 0.93 & 0.66 & System vs. Expert1\\
	
		\hline
		\multirow{3}{*}{\textbf{6 $\times$ 6 millimeters}} & 0.68 & 0.74 & Expert1 vs. Expert2\\
																							& 0.48 & 0.66 & System vs. Expert2\\
																							& 0.40 & 0.72 & System vs. Expert1\\
		\hline
 \end{tabular}
\end{center}
\end{adjustwidth}
\end{table}

On the other hand, the obtained correlation coefficients with the diabetic image subset are presented in Table \ref{tab:corr_coef_diabetic}. 

\begin{table}[ht!]
\begin{adjustwidth}{-2.25in}{0in} 
\caption{\bf Correlation coefficients that were obtained using the manual and the automatic area size measurements in diabetic OCT-A images.}
\label{tab:corr_coef_diabetic}
\begin{center}
 \begin{tabular}{|l+l|l|}\hline
		\textbf{Size} & \textbf{Superficial} & \textbf{Deep}\\ \thickhline
		\bf $3 \times 3$ millimeters & 0.93 & 0.92 \\ \hline
		\bf $6 \times 6$ millimeters & 0.71 & 0.49 \\ \hline
		
 \end{tabular}
\end{center}
\end{adjustwidth}
\end{table}

Despite the satisfactory results of the area size correlations, we analysed not only the final measured area sizes but also the specific matching degree of both extracted regions. In that sense, we performed an additional analysis of the manual and the computational segmented FAZ regions using the Jaccard's index \cite{10.2307/2413572,jaccard_index_segmentation_image}. We used this index given its simplicity and accurate representation of the agreement degree, frequently used in a large variability of domains and, specifically, in the evaluation of medical image segmentation issues \cite{Bouix07,Silva11,Lassen15,gonccalves2016hessian}. The Jaccard's index is defined by:

\begin{equation}
Jaccard = \frac{A \cap B}{A \cup B}
\label{Eq:Jaccard}
\end{equation}

where $A$ and $B$ represent the regions of the segmentations that are compared. The Jaccard's index tends to one with high levels of agreement. In this case, with largely similar segmentations, their intersection is practically the same as their union. On the contrary, the Jaccard's index tends to zero for a reduced level of agreement. The Jaccard's index presents values in the range $ [0,1] $, being generally considered the obtained values as:

\begin{itemize}
\item \textbf{Poor}. If the Jaccard's index is $ 0.4 $ or less, it is considered a poor result.
\item \textbf{Good}. If the obtained value with the Jaccard's index is approximated to $ 0.7 $ the result is considered good.
\item \textbf{Excellent}. In the case that the Jaccard's index takes values of $ 0.9 $ or higher, the result of the segmentation is considered excellent.
\end{itemize}


Table \ref{tab:jaccard01} details the average Jaccard's indexes that were obtained for all the analysed images using the manual and the automatic segmented regions in healthy OCT-A images. The results were divided in 4 parts using both size and depth dimensions, as mentioned, with the typical configurations that the specialists normally use: superficial \& 3x3 millimeters, superficial \& 6x6 millimeters, deep \& 3x3 millimeters and deep \& 6x6 millimeters. We divided the analysis in this 4 subgroups given that each case presents specific characteristics and complexity obtaining, therefore, a more adjusted analysis of the performance of the method. In addition, there were divided into other 3 subgroups, based on the comparisons that were performed (comparison between both experts or between each expert and the automatic segmentation). In general terms, we can see that all the cases reached satisfactory results, but presenting slight variations that are discussed in detail next.

\begin{figure}
\centering
\includegraphics[width=120mm]{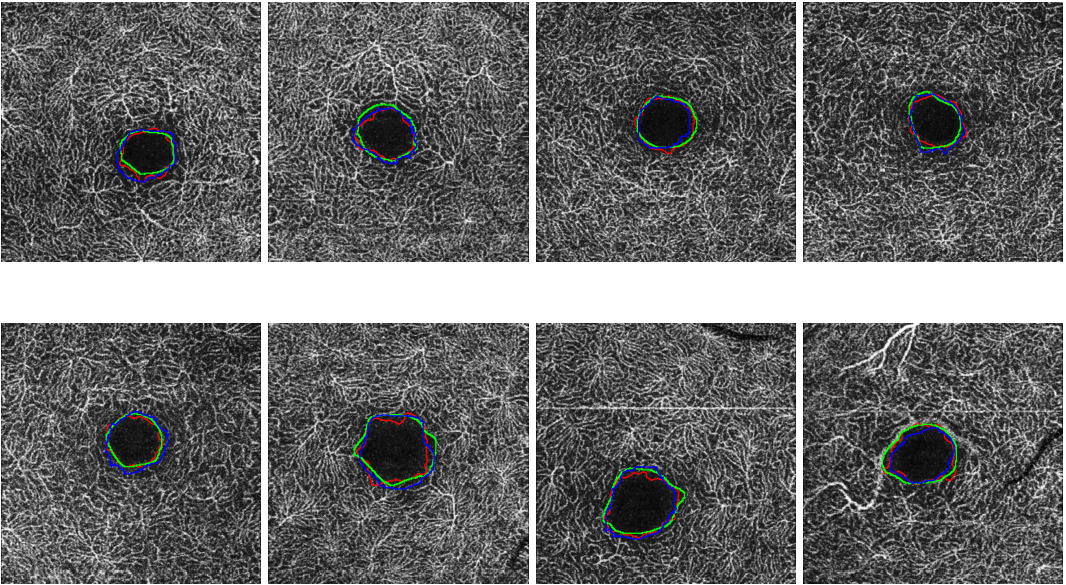}

	\caption{Comparative examples with goods and bad results in the Jaccard's index in the four subgroups (superficial and deep in $3 \times 3$ and $6 \times 6$ sizes).}
	\label{fig:jaccard_best_and_worst}
\end{figure}

\begin{table}[ht!]
\begin{adjustwidth}{-2.25in}{0in}
\caption{\bf Jaccard indexes that were obtained for each subgroup of healthy OCT-A images.}
\label{tab:jaccard01}
\begin{center}
 \begin{tabular}{|l+l|l|l|}\hline
		\textbf{Size} & \textbf{Superficial} & \textbf{Deep} & \textbf{Comparisons}\\
		\thickhline
		\multirow{3}{*}{\textbf{3 $\times$ 3 millimeters}} & 0.83 & 0.72 & Expert1 vs. Expert2\\
																							& 0.82 & 0.72 & System vs. Expert2\\
																							& 0.81 & 0.74 & System vs. Expert1\\
	
		\hline
		\multirow{3}{*}{\textbf{6 $\times$ 6 millimeters}} & 0.72 & 0.68 & Expert1 vs. Expert2\\
																							& 0.77 & 0.72 & System vs. Expert2\\
																							& 0.72 & 0.66 & System vs. Expert1\\
		\hline
 \end{tabular}
\end{center}
\end{adjustwidth}
\end{table}
Images with a size of 6x6 millimeters typically present smaller FAZ regions, which means that small variations and imperfections in the segmentation process of the system and/or the expert impact and penalize in a higher rate the obtained agreements, producing lower Jaccard's indexes than the results with 3x3 millimeters images that include a zoom with more resolution of the FAZ region. In addition, deep images (as stated above) present more diffuse, small and rough edges which constitutes a more complex scenario. In these cases, the computational results are slightly more irregular, given that they are based in the intensity characteristics, than the manual labelling given the expert tried to produce a smoother manual segmentation.
Given that, the Jaccard's indexes were slightly penalized, although in the graphic results (examples are presented in Fig. \ref{fig:jaccard_best_and_worst}) we can appreciate similar results and, even in this case, Jaccard's indexes approximate values of $0.7$, which are considered satisfactory. In addition, the Jaccard's index between the specialists is, in all the four cases, similar to the Jaccard's index between the system and both the experts. Therefore, we consider that the automatic segmentation is satisfactory in relation with to the results of both specialist.

Table \ref{tab:jaccard02} details the average Jaccard's index using the expert's annotations and the system's extracted region in the different image subgroups. As in the previous case, the $3 \times 3$ superficial case represents the subgroup with the highest results whereas the $6 \times 6$ deep case provided the lowest values, as we explained before. 

\begin{table}[ht!]
\begin{adjustwidth}{-2.25in}{0in}
\caption{\bf Jaccard's indexes that were obtained for each subgroup of diabetic OCT-A images.}
\label{tab:jaccard02}
\begin{center}
 \begin{tabular}{|l+l|l|}\hline
		\textbf{Size} & \textbf{Superficial} & \textbf{Deep} \\ \hline
		\textbf{$3 \times 3$ millimeters} & 0.84 & 0.76 \\ \hline
		\textbf{$6  \times 6$ millimeters} & 0.68 & 0.68 \\ \hline
 \end{tabular}
\end{center}
\end{adjustwidth}
\end{table}

\begin{figure}[H]
\centering
\includegraphics[width=120mm]{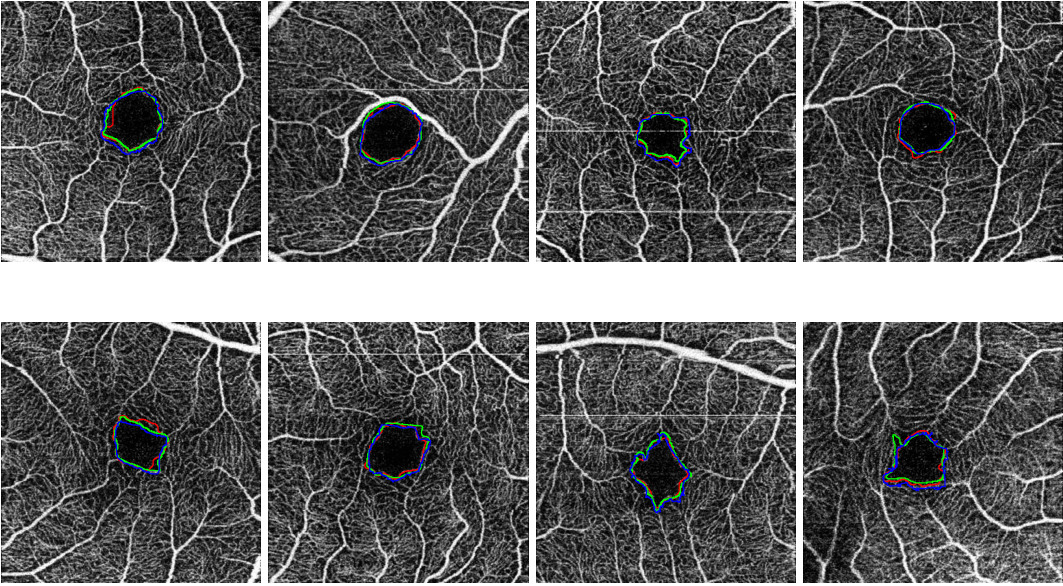}
  
	\caption{Comparative examples of the experts (green and red) and the automatic computational (blue) FAZ measurements in superficial 3 millimeters images.}
	\label{fig:jaccardsup3}
\end{figure}

The case that presented the best results was the one including OCT-A images with a size of 3x3 millimeters and at a superficial depth (see Fig. \ref{fig:jaccardsup3}), given they present FAZ regions with better marked contours and larger sizes. The size of the FAZ region influences the Jaccard index since the greater that is the size, the less that it is penalize by variations in the contour.

\begin{figure}[H]
\centering
\includegraphics[width=120mm]{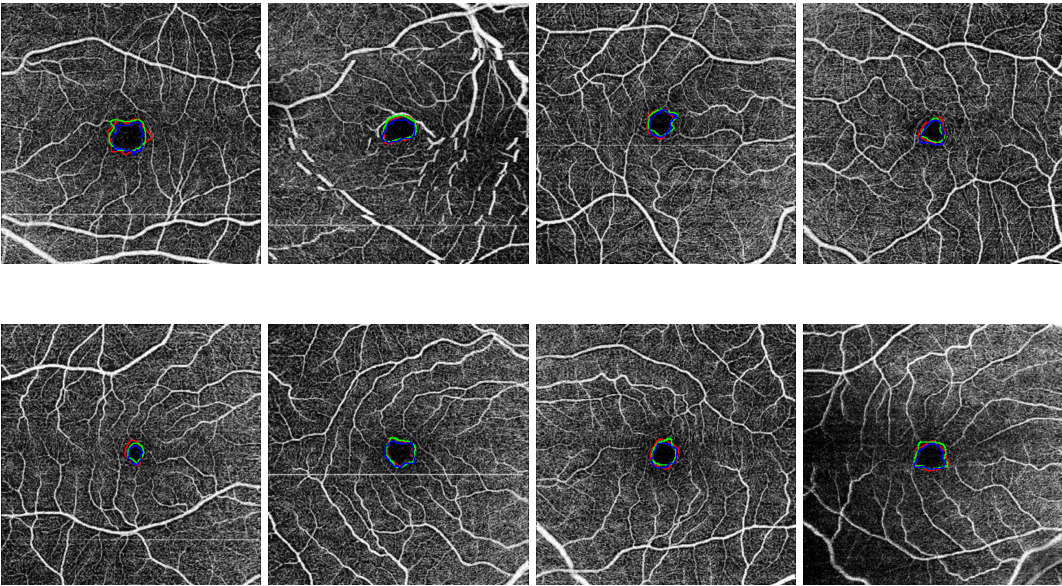}
  
	\caption{Comparative examples of the experts (green and red) and the automatic computational (blue) FAZ measurements in superficial 6 millimeters images.}
	\label{fig:jaccardsup6}
\end{figure}

Images with a size of 6x6 millimeters and at a superficial depth (see Fig. \ref{fig:jaccardsup6}), also provided satisfactory results, having clear and marked edges, allowing that the segmentation of the system and the labelling of the expert are significantly similar. Despite that, given that the FAZ region is smaller, changes that are generated in the contour identification affect in a higher rate the Jaccard's index.

\begin{figure}[H]
\centering
\includegraphics[width=120mm]{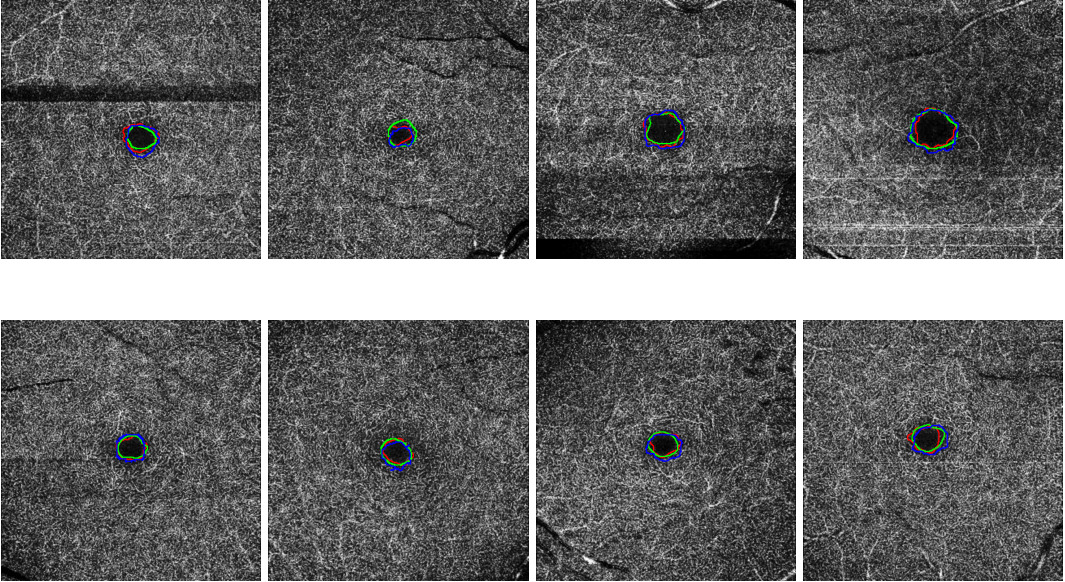}
  
	\caption{Comparative examples of the experts (green and red) and the automatic computational (blue) FAZ measurements in deep 3 millimeters images.}
	\label{fig:jaccardprof3}
\end{figure}

\begin{figure}[H]
\centering
\includegraphics[width=120mm]{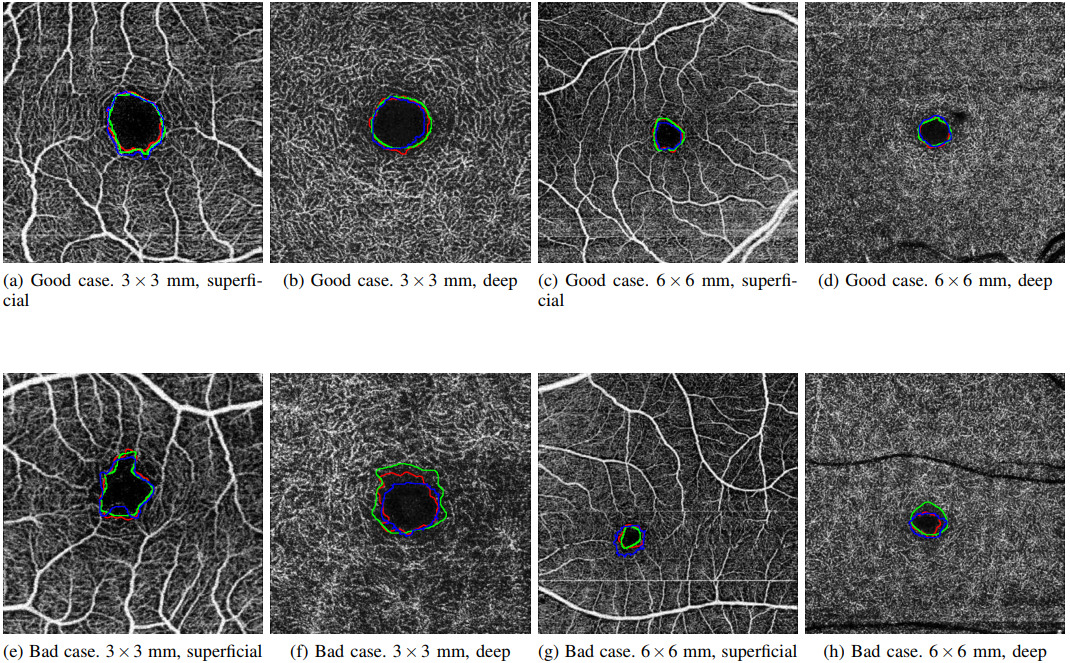}
  
	\caption{Comparative examples of the experts (green and red) and the automatic computational (blue) FAZ measurements in deep 6 millimeters images.}
	\label{fig:jaccardprof6}
\end{figure}

Finally, deep images (see Fig. \ref{fig:jaccardprof3} and \ref{fig:jaccardprof6}) are those that retrieved the worse results, nevertheless they remain within the range of Jaccard index results that are considered correct. From both sizes coherently once again 3x3 millimeters images presented better results.

Additionally, Fig. \ref{fig:jaccard_best_and_worst} presents examples of the best and worst cases of each subgroup, demonstrating that frequently even in the worst scenario, the method provides acceptable results in the segmentation of the FAZ region and the calculation of the corresponding area size for the following clinical analysis. In addition, the summary of all the subgroups of Jaccard's index is represented in the Table \ref{tab:jaccard_total}, where there are the best and worst cases of this metric in each subgroup.
 
\begin{table}[ht!]
\begin{adjustwidth}{-2.25in}{0in}
\caption{\bf Worst and best Jaccard's indexes that were obtained for each subgroup of OCT-A images.}
\label{tab:jaccard_total}
\begin{center}
 \begin{tabular}{|l+l|l|l|}\hline
		\textbf{Size} & \textbf{Superficial} & \textbf{Deep} & \textbf{Comparisons}\\
		\thickhline
		\multirow{3}{*}{\textbf{3 $\times$ 3 millimeters best}} & 0.92 & 0.83 & Expert1 vs. Expert2\\
																									 & 0.93 & 0.88 & System vs. Expert2\\
																							     & 0.88 & 0.86 & System vs. Expert1\\
		\hline
		\multirow{3}{*}{\textbf{3 $\times$ 3 millimeters worst}} & 0.74 & 0.44 & Expert1 vs. Expert2\\
																							      & 0.68 & 0.59 & System vs. Expert2\\
																							      & 0.69 & 0.57 & System vs. Expert1\\
	
		\hline
		\multirow{3}{*}{\textbf{6 $\times$ 6 millimeters best}} & 0.84 & 0.84 & Expert1 vs. Expert2\\
																							     & 0.88 & 0.87 & System vs. Expert2\\
																							     & 0.88 & 0.83 & System vs. Expert1\\
		\hline
		\multirow{3}{*}{\textbf{6 $\times$ 6 millimeters worst}} & 0.43 & 0.40 & Expert1 vs. Expert2\\
																							      & 0.43 & 0.40 & System vs. Expert2\\
																							      & 0.42 & 0.41 & System vs. Expert1\\
		\hline
 \end{tabular}
\end{center}
\end{adjustwidth}
\end{table}

\section*{Discussion and Conclusions}
\label{sec:conclusions}

There exist many vascular diseases that affect the retinal micro circulation, not only specific vascular diseases of the human eye but also others of general impact in the patients, as hypertension or diabetes. For that reason, the availability of automatic tools that quickly calculate suitable biomarkers and assist clinicians in the diagnosis and monitoring of patients is of great interest in the healthcare systems.

Among the different ophthalmological image modalities, we can find the recent appearance of the OCT-A image modality, offering visualizations of the characteristics of the retinal vasculature at different depths, but being non-invasive as it omits the injection of fluorescein, as happens with the case of classical angiographies. Given its utility, the OCT-A image modality is increasing its interest in clinical and research practise. The automatic extraction of the FAZ region in OCT-A images is of a great interest given it offers important advantages in many aspects with respect to the manual extraction of the specialist. In addition to the avoidance of a tedious manual labelling process with an computational and instant tool, the automatic extraction provides repeatability and determinism, which is largely complicated with the manual extractions of the clinical experts, representing a fundamental characteristic in accurate diagnostic and monitoring processes. 

In this work, we present a novel automatic methodology that identifies and precisely segments the FAZ region using OCT-A images. The proposed method applies morphological operators to enhance the vascular brightness of the OCT-A images. Subsequently, edge detection techniques are performed to eliminate unnecessary spurious details and detect the vascular regions. After this, morphological operations are performed, again, to eliminate areas that are not of interest in the detection of the aimed FAZ region and keep a reduced number of candidates. Then, specific domain knowledge is used to preserve, from all the candidates, the most suitable identification as the FAZ localization. Finally, a region growing approach is applied using this preliminary identification as seed to obtain a precise segmentation as the final FAZ segmentation result. Additionally, using this precise segmentation, the method calculates the corresponding FAZ area size, as an important biomarker for its use in the study of the evolution of different relevant diseases and their treatments.

Regarding the obtained results with the used image dataset, the FAZ localization obtained a success rate over a 97\%, as well as a correlation coefficient about a 0.9 in $3 \times 3$ superficial images (better case),  whereas a coefficient of 0.7 in $6 \times 6$ deep images (worse case), using the manual performance of the clinical experts as reference. The similarity results were also measured with the Jaccard's index, obtaining an average value of 0.8 in 3x3 millimeters superficial images (better case) and a 0.7 of average value in 6x6 millimeters deep images (worse case). Summarizing, we can conclude that the proposed method offered a satisfactory performance in all the designed scenarios.

To perform the validation process, we tested the method with the public image dataset \textit{OCTAGON} \cite{octagon}, that contains 168 images grouped in 2 image subsets: the first one, formed by 144 healthy OCT-A images; and the second one, formed by 24 diabetic OCT-A images. The healthy dataset is divided into different groups of ages (10-19, 20-29, 30-39, 40-49, 50-59 and 60-69 years old) with 3 patients in each age-range. Each of these patients contains OCT-A images from each eye (left and right), containing both of them one image of each subgroup ($3 \times 3$ millimeters in superficial, $3 \times 3$ millimeters in depth, $6 \times 6$ millimeters in superficial and $6 \times 6$ millimeters in depth).
The healthy image subset also provides the manual labelling of 2 experts, allowing us to proceed with a robust validation. On the other hand, the diabetic subset contains 24 images, 6 of each subgroup ($3 \times 3$ millimeters in superficial, $3 \times 3$ millimeters in depth, $6 \times 6$ millimeters in superficial and $6 \times 6$ millimeters in depth). This subset also contains the manual labeling of an expert clinician.
As we can see, we use a complete dataset that contains healthy and pathological images, with a large variability of OCT-A images in the different indicated age-ranges  (specially in the healthy case) also including, at least, manual annotations of one expert clinician, as said allowing us a robust validation. The different methods of the state of the art faced only datasets with 1 or 2 of the subgroups that we use in our proposal, as we can see in Table \ref{tab:comparative01}, where we compare the coverage of our OCT-A image dataset with different published works.

\begin{table}[ht!]
\begin{adjustwidth}{-2.25in}{0in}
\caption{\bf Comparative of the coverage OCT-A image types between this proposal and the Lu \textit{et al.} \cite{lu2018evaluation} and Hwang \textit{et al.} \cite{article} works.}
\label{tab:comparative01}
\begin{center}
 \begin{tabular}{|l+l|l|l|}\hline
		\textbf{Method} & \textbf{Superficial} & \textbf{Deep} & \textbf{Size} \\ \thickhline
		\multirow{2}{*}{\textbf{Proposed}} & $\surd$ & $\surd$ & $3 \times 3$ millimeters \\
		& $\surd$  & $\surd$ & $6 \times 6$ millimeters\\\hline
																			
		\multirow{2}{*}{\textbf{Lu \textit{et al.}\cite{lu2018evaluation}}} & $\surd$ & $-$ & $3 \times 3$ millimeters \\
		& $-$  & $-$ & $6 \times 6$ millimeters\\\hline
																																				
		\multirow{2}{*}{\textbf{Hwang \textit{et al.}\cite{article}}} & $-$  & $-$ & $3 \times 3$ millimeters \\
		& $\surd$ & $-$ & $6 \times 6$ millimeters\\\hline

 \end{tabular}
\end{center}
\end{adjustwidth}
\end{table}

Regarding the results, we tested the performance of the proposed method using the correlation coefficient of Pearson and the Jaccard's index. The first one is useful to prove that the manual and the automated extracted areas are related. The second one is useful to check the coverage area between the manual and automated extractions as well.
As we saw in the Section \textit{\nameref{e2}}, the results in both validation test are satisfactory, concluding that the method correlates accurately with the manual labelling of the experts. 
To compare our approach with other similar works, we can see Table \ref{tab:compLu}. There, we can check the results of the Jaccard's index in healthy and diabetic cases with our proposal and the Lu \textit{et al.} \cite{lu2018evaluation} method.
Given that our image dataset fits better with the real conditions that are faced by the expert clinicians, including a significative variability in the image conditions, as detailed, we implemented a more general solution than other proposals. For this reason, our method provided slightly lower results in $3 \times 3$ millimeters, superficial, representing in any case satisfactory results.
In fact, we obtain satisfactory results in all the subgroups that were tested, both in healthy and also in diabetic OCT-A images. In this comparative, no results were reported regarding the work of Hwang \textit{et al.}\cite{article} given that their proposal is centered in clinical research, and being postponed the validation of the method as future work.

\begin{table}[ht!]
\begin{adjustwidth}{-2.25in}{0in}
\caption{\bf Comparison of the Jaccard's indexes that were obtained for $3 \times 3$ millimeters superficial OCT-A images in our proposal and Lu \textit{et al.} method \cite{lu2018evaluation}.}
\label{tab:compLu}
\begin{center}
 \begin{tabular}{|l+l|l|}\hline
		\textbf{Method} & \bf Proposed & \bf Lu \textit{et al.} \cite{lu2018evaluation} \\ \thickhline
		\textbf{Superficial $3 \times 3$ mm healthy case} & 0.82 & 0.87 \\
		\textbf{Superficial $3 \times 3$ mm diabetic case} & 0.84 & 0.82 \\\hline
 \end{tabular}
\end{center}
\end{adjustwidth}
\end{table}

To further test the robustness and suitability of the obtained results, it is proposed as future work to design experiments that involve image datasets of patients with different relevant pathologies that affect the retinal vascularity. On the other hand, it is proposed the use of the proposed methodology to perform the measurements of the FAZ region in real scenarios, monitoring pathologies to confirm the validity of the method.
All the code developed in this work is publicly available on the repository\footnote{https://github.com/macarenadiaz/FAZ\_Extraction}.

\section*{Acknowledgments}

This  work  has  received  financial  support  from  the  European Union  (European  Regional  Development  Fund  -  ERDF)  and  the  Xunta  de Galicia, Centro singular de investigación de Galicia accreditation 2016-2019, Ref. ED431G/01; and Grupos de Referencia Competitiva, Ref. ED431C 2016-047.


%
%
%

\end{document}